\documentclass{article}

\usepackage{PRIMEarxiv}

\usepackage[utf8]{inputenc} % allow utf-8 input
\usepackage[T1]{fontenc}    % use 8-bit T1 fonts
\usepackage{hyperref}       % hyperlinks
\usepackage{url}            % simple URL typesetting
\usepackage{booktabs}       % professional-quality tables
\usepackage{amsfonts}       % blackboard math symbols
\usepackage{nicefrac}       % compact symbols for 1/2, etc.
\usepackage{microtype}      % microtypography
\usepackage{lipsum}
\usepackage{fancyhdr}       % header
\usepackage{graphicx}       % graphics
\usepackage[table]{xcolor}
\usepackage{multirow}
\usepackage{makecell}

\graphicspath{{media/}}     % organize your images and other figures under media/ folder

\title{Brain-Inspired Deep Imitation Learning for Autonomous Driving Systems}
\author{
  Hasan Bayarov Ahmedov \\
  Department of Computing Science \\
  University of Aberdeen\\
   King's College, Aberdeen,\\
                    AB24 3FX\\
  \texttt{hasan.ahmedov.17@abdn.ac.uk} \\
  \And
  Dewei Yi\\
  Department of Computing Science \\
  University of Aberdeen\\
   King's College, Aberdeen,\\
                    AB24 3FX\\
  \texttt{dewei.yi@abdn.ac.uk} \\
   \And
   Jie Sui \\
  School of Psychology \\
                    University of Aberdeen\\
                    King's College, Aberdeen,\\
                    AB24 3FX \\
  \texttt{jie.sui@abdn.ac.uk} \\
}

\begin{document}
\maketitle

\begin{abstract}
    Autonomous driving has attracted great attention from both academics and industries. To realise autonomous driving, Deep Imitation Learning (DIL) is treated as one of the most promising solutions, because it improves autonomous driving systems by automatically learning a complex mapping from human driving data, compared to manually designing the driving policy. However, existing DIL methods cannot generalise well across domains, that is, a network trained on the data of source domain gives rise to poor generalisation on the data of target domain. In the present study, we propose a novel brain-inspired deep imitation method that builds on the evidence from human brain functions, to improve the generalisation ability of deep neural networks so that autonomous driving systems can perform well in various scenarios. Specifically, humans have a strong generalisation ability which is beneficial from the structural and functional asymmetry of the two sides of the brain. Here, we design dual Neural Circuit Policy (NCP) architectures in deep neural networks based on the asymmetry of human neural networks. Experimental results demonstrate that our brain-inspired method outperforms existing methods regarding generalisation when dealing with unseen data. Our source codes and pretrained models are available at \href{https://github.com/Intenzo21/Brain-Inspired-Deep-Imitation-Learning-for-Autonomous-Driving-Systems}{https://github.com/Intenzo21/Brain-Inspired-Deep-Imitation-Learning-for-Autonomous-Driving-Systems}.
\end{abstract}

% keywords can be removed

\section{Introduction}
Autonomous vehicles have received the attention of various industries and academic institutions. Due to their potential to improve the safety and efficiency of the driving experience, they are expected to have a massive economic impact within the next decade \cite{kim2009curve}. A machine learning (ML) approach, known as end-to-end learning, has been used to achieve such an autonomous driving system \cite{bojarski}. This approach refers to the use of a single, self-contained system that automatically translates a sensory input, such as a captured image, into a set of instructions for driving. This type of learning is called Deep Imitation Learning (DIL) \cite{laskey, kumaar2018}. 

A number of DIL algorithms have been developed in last two decades that overcome the lack of vast data for training procedures in Deep Learning (DL) models. For example, a data-driven, off-policy IL learning called Behavioural Cloning (BC) has been proposed to simulate human driving \cite{d2003}, which replaces deterministic behaviour  with  a  human-driven  paradigm that is sophisticated  and  flexible  in  its  response  to different visual environments \cite{saksena2019}. This method also allows a final system to be built by improving its ability to generalise, that is, enabling the model to operate in a variety of circumstances \cite{chen2015}.

Successful deep neural networks with a single, task-specific algorithm should be able to generalise well across domains. Such an algorithm can enable simultaneous expression of generalisation ability by learning coherent representations and interpretable dynamics explanations of the world \cite{lecun}. To some extent, deep neural networks, e.g., CNN, have shown generalisation ability on image classification, recognition \cite{krizhevsky} and autonomous driving/collision avoidance \cite{bojarski2017, kumaar} problems. However, this ability still needs to be further improved, especially in the field of autonomous driving, which requires reflecting a spatiotemporal understanding of dynamic driving settings. From this point onwards, the primary aim of the present study is to explore whether a certain architecture (built on evidence in the human brain) generalises better than other architectures in autonomous driving models, by examining a range of methodologies, including Convolutional Neural Networks, single neural circuit policy networks, and dual neural circuit policy networks. We evaluate generlisation performance on diverse datasets, such as Udacity simulator dataset and comma.ai dataset. 

In the present study, we employ brain-inspired architectures into deep neural networks, where we build various end-to-end DIL model architectures and put them into competition with the reference neural network mentioned in \cite{bojarski}. These network architectures are compared in terms of the training and generalisation Mean Squared Error (MSE) result between the model predictions and the actual steering angles provided by a human driver. The present brain-inspired models are built on the evidence of the asymmetry of neural connectivity in the human brain, because converging evidence from structural and functional neuroimaging studies has shown that the neural basis of human cognitive functions is extensively distributed in different regions across the two sides of the brain, which indicates the complexity of how humans respond to the environment via different levels of processing \cite{brancucci2009asymmetries}. From an evolutionary perspective, the lateralised functions of human two brains increases with evolution \cite{duboc2015asymmetry}. Specifically, the present study adopts DIL in accomplishing dynamic vehicle control (e.g. steering angle, speed). A Udacity simulator \cite{udacitysim} has been employed as a starting point which has been later replaced by the real-world comma.ai dataset \cite{commares}. Attempting to solve the task has required  the  utilisation  of  the  Neural Circuit Policies (NCPs)  model, a designed sparse RNN based on the Liquid Time Constant (LTC) neuron with synapse model, which is inspired by C. elegans organism nervous system \cite{ncpgit}. Therefore, the key contribution of our work is summarised as follows.
    \begin{itemize}
    
        \item We propose a brain-inspired DIL method to enhance the generalisation ability of autonomous driving models, where the design of dual NCP architectures replicates the human brain left and right hemispheres.
        
        \item Various modifications of the newly developed composite networks are explored for fully leveraging the strengths of brain-inspired network architectures.
        
        \item Our method is evaluated on the Udacity simulator and comma.ai datasets to demonstrate its performance against contemporary architectures.
    \end{itemize}

\section{Related Work}

\subsection{Convolutional Neural and Recurrent Neural Networks}

    The Convolutional Neural Networks (CNNs) are one of the most popular type of Deep Neural Network (DNN) \cite{bonin2008visual}. It is named after the mathematical linear operation between matrices called convolution. This model has yielded groundbreaking results in a variety of pattern recognition fields, spanning from image processing to speech recognition, in the last years. Furthermore, it has been shown that CNNs perform exceptionally when it comes to ML problems. In particular, these networks are implemented in image data challenges. For instance, CNNs have achieved outstanding results on the largest image classification dataset named ImageNet, Computer Vision (CV) problems, and on Natural Language Processing (NLP).
    
%\subsection{Recurrent Neural Networks}

% https://icml.cc/2011/papers/524_icmlpaper.pdf

% https://www.ibm.com/cloud/learn/recurrent-neural-networks

Recurrent neural networks (RNNs) are a type of networks which are able to extract useful from sequential or time series data. RNNs show impressive performance on ordinal or temporal problems, such as language modelling \cite{mikolov} and translation, NLP, speech recognition \cite{graves}, machine translation \cite{kalchbrenner} and image captioning. They are integrated into many well-known applications e.g. Siri, voice search, and Google Translate etc. Similar to Feed Forward Neural Networks (FNNs) and CNNs, RNNs utilise training data to learn. RNNs were developed in response to a number of flaws in the FNN model \cite{simplilearn}: inability to handle sequential data; managing only the current input; no previous input memorisation.

\subsection{Neural Circuit Policies}
\begin{figure}
    \centering
    \begin{tabular}{c}
    \frame{\includegraphics[width=.95\linewidth]{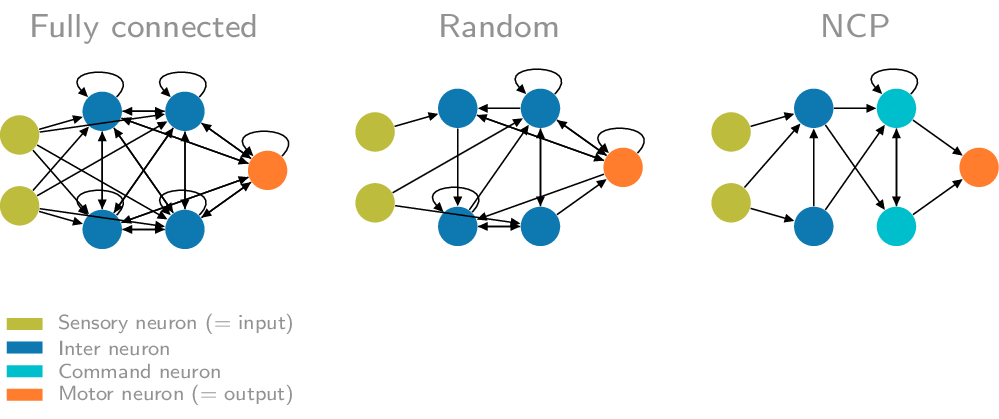}}
    \end{tabular}
    \caption{FC vs. random vs. NCP wirings \cite{ncpgit}.}
    \label{fig:wirings}
    \end{figure}

    The architecture of NCPs (Figure \ref{fig:wirings}) is galvanised by the C. elegans nematode wiring representation \cite{wicks}. The wiring diagram of C. elegans attains a sparsity close to 90\% \cite{yan2017} primarily consisting of feed forward links from sensors to intermediate neurons. Also, it comprises exceedingly cyclical connections between inter and command neurons and feed forward links from command to motor neurons. This particular topology has been shown to have computational benefits such as effective distributed control with a minimal number of neurons \cite{yan2017}, hierarchical temporal dynamics \cite{kaplan2020}, robot-learning ability \cite{lechner2019}, and maximal knowledge transmission in sparse-flow networks \cite{hasani2020}. The neural dynamics of NCPs are defined by continuous-time Ordinary Differential Equations (ODEs), which have been originally designed to model the nervous system dynamics of small beings like C. elegans \cite{hasaniltc}. At the heart of NCPs stands a nonlinear time-varying synaptic propagation system, which enhances their expressiveness in modelling time series as opposed to their DL counterparts \cite{hasaniltc}. Moreover, LTCs are the foundational neural building blocks of NCPs \cite{hasaniltc}.
    
    Brain-inspired NCP models paired with compact CNNs \cite{lecun90} have been shown promising performance on learning how to drive a vehicle directly from high-dimensional inputs in very minimal networks (19 neurons). NCPs are also advantageously employed in full-stack autonomous vehicle systems \cite{lechner}. These neural network agents are capable of driving a vehicle on previously unknown routes while being resistant to input artefacts, acquiring short-term causal concepts, and achieving comprehensible dynamics. 

%------------------------------------------------------------------------- 
\subsection{Deep Learning in Autonomous Vehicles}
    The behavioural cloning related autonomous driving method was pioneered back in the late '80s by Autonomous Land Vehicle In a Neural Network (ALVINN) \cite{pomerleau}. ALVINN is described as a 3-layer back-propagation network %, illustrated in Figure \ref{fig:alvinn},
    designed for road following. ALVINN's architecture is straightforward, consisting only of a fully connected network, which is relatively tiny by contemporary standards. Even though the network has been applied to uninvolved navigation environments, it shows how neural architectures can be used for end-to-end autonomous navigation. A similar approach was taken later in 2005 to train, this time, a CNN to drive an off-road mobile robot \cite{muller}. More recently, Bojarski et al. \cite{bojarski} propose a deeper CNN structure for lane following based solely on a front-facing camera. In all these cases, a DNN is found to be surprisingly effective at learning a complex mapping from a raw image to vehicle control.

    %%%% https://ieeexplore.ieee.org/stamp/stamp.jsp?tp=&arnumber=8675582
    Recent research activities have focused on employing neural structures in grasping control policies and behaviours from raw sensor data.  The majority of these approaches are accentuate either on Reinforcement Learning (RL) \cite{zhu2017} or Supervised Learning (SL) \cite{smolyanskiy2017, du2019}. Furthermore, intriguing autonomous navigation architectures for self-driving cars \cite{woo2017} and event-based vision \cite{maqueda2018} for steering angle prediction have been suggested. However, none of these techniques has regarded the temporal domain for autonomous navigation.Blended neural networks, which fuse the advantages of many forms of neural networks, are gaining popularity in the field of CV \cite{lin2017}. Also, these hybrid architectures have been proficiently administered and signified their performance in image captioning \cite{donahue2015}, image processing \cite{wang2016}, action recognition \cite{ballas2015}, and other applications. Regardless, composite neural structures' appliance in time series analysis is rather restricted. Combining CNN and Long Short-Term Memory (LSTM) can be done in a variety of ways, including training the models individually and then combining them, or the other way around \cite{deng2014}.  The CNN's convolutions can be used explicitly for reading data into the LSTM modules and such a stacked neural network configuration can  be  referred  to  as  convolutional LSTM or CNN-LSTM for short \cite{shi}. Du et al. \cite{du2017} suggests an architecture for traffic flow forecasting that is derived from LSTM and CNN hybridisation. In traffic flow results, the structure emphasises the effects of local spatial and long dependency functions, as well as spatiotemporal associations. Lin et al. \cite{lin2017} propose an end-to-end hybrid neural network (CNN and LSTM) for studying local and global contextual features for time series pattern prediction. CNN and LSTM complement each other with the strengths of CNN in extracting spatial information and LSTM in temporal information \cite{sainath}.

\section{Brain-inspired Deep Imitation Learning}
    Various brain-inspired deep network networks are created to imitate human driving behaviour and resolve the sophistication of autonomous lane-keeping by comparing them and finding the best performing model. Initially, a CNN network is built reckon with the Nvidia's framework as described in the paper of Bojarski et al. \cite{bojarski}, which has been demonstrated as effective approach in the field of autonomous vehicles. In our work, this network is used as a feature extractor located at front-end of the stacked models. We add the a bio-inspired RNN structure, called NCP, to the CNN head for supporting the brain-inspired frameworks in sequential information extraction and performing vehicle dynamics control. 
    
    All of the developed models are trained by end-to-end learning, which directly links raw pixels from a single dash cam to steering instructions. In this paper, we use the autonomous lane following as the main criteria for assessing the performance of the developed models. We train the weights of our networks by minimising the MSE between the model predicted steering angles and the ground truth of steering angles. In addition, we employ the Udacity simulator environment to better visualise our models' generalisation findings. This is accomplished by running the software to autonomously drive the automobile on both of the specified tracks using the pretrained architectures. The details of feature extractor (CNN model architecture) and brain-inspired network architectures used in our work are discussed as follows.
    
    \subsection{CNN Model Architecture}
\begin{figure}[h]
    \centering
    \begin{tabular}{c}
    \frame{\includegraphics[width=10cm, height=6cm]{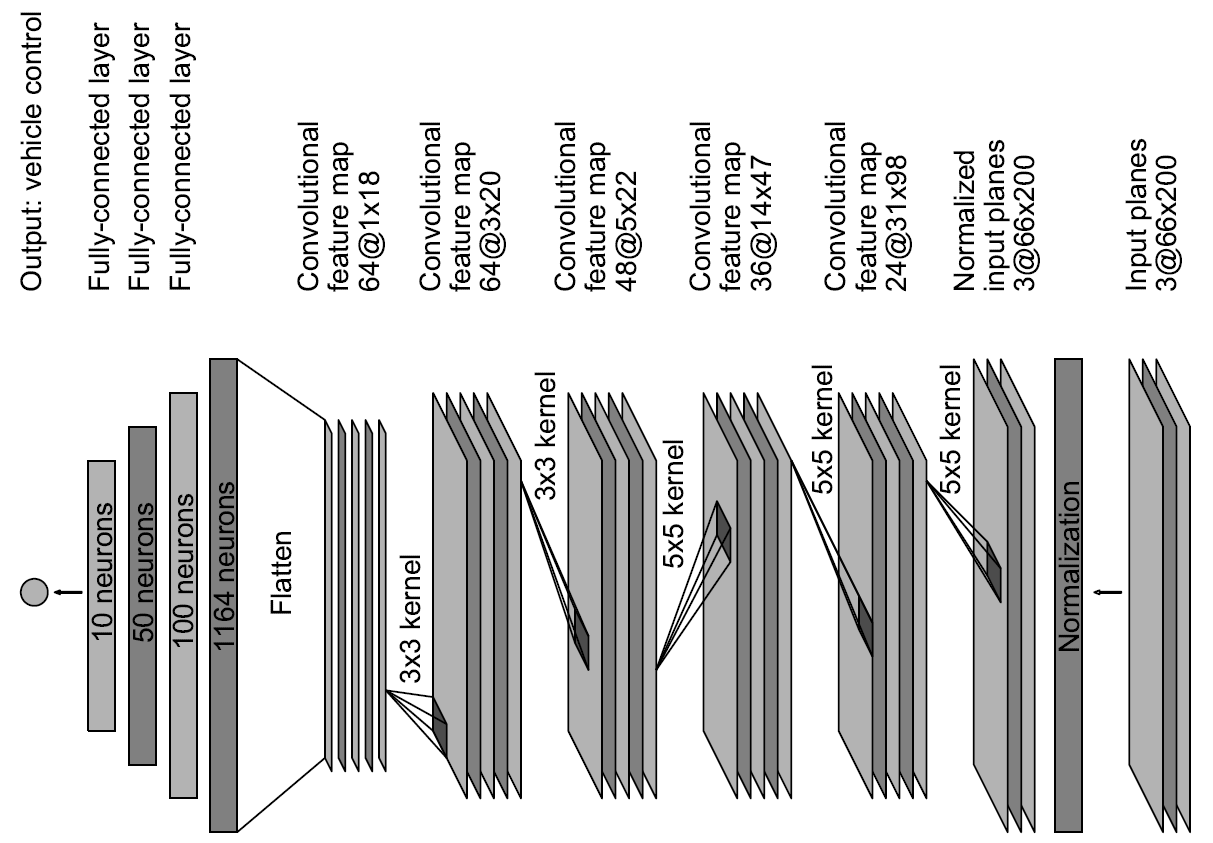}}
    \end{tabular}
    \caption{Nvidia CNN structure \cite{bojarski}.}
    \label{fig:cnn}
    \end{figure}

    The CNN model used in Figure \ref{fig:cnn} consists of 9 layers, including a lambda normalisation layer, 5 Convolutional (Conv2D) layers, and 3 Fully Connected (FC/Dense) layers. For 5 Convolutional layers, the first three Convolutional layers are with 5-by-5 kernel and last two layers Convolutional are with 3-by-3 kernel. The three FC layers are with 1164, 100, 50, and 10 neurons, respectively. The CNN model is used as a baseline model in this paper.
    
    Since the CNN model used in our brain-inspired framework is to perform as a spatial feature extractor, we only need the its front layers. To this end, the first six layers are preserved and concatenated with the brain-inspired layers. In addition, a dropout layer is added for adapting to brain-inspired framework.
    
    \subsection{Brain-inspired Network Architectures}
        
        Brain-inspired network architectures stack the front-end layers of above mentioned CNN model  with a single and dual NCP-based RNN structures, which are named CNN-NCP and CNN-DNCP accordingly. Prior to determining the most promising combination of CNN and NCP models, the evaluations of various combinations are carried out to find out the best one, where the different settings of the Dense layers of the CNN and their units are evaluated. To make different method more comparable with others, we attempt to make the resultant versions to be as identical as possible to the original one. Considering the performance results of the evaluated adjustments, a Dense layer with 100 units is chosen.  After the last convolutional layer, we apply the aforementioned per-channel linear Dense layer to obtain 100 latent features serving as sensory inputs to the RNN compartment. 
        
\begin{figure} [h]
        \centering
        \begin{tabular}{cc}
        \frame{\includegraphics[width=.75\linewidth]{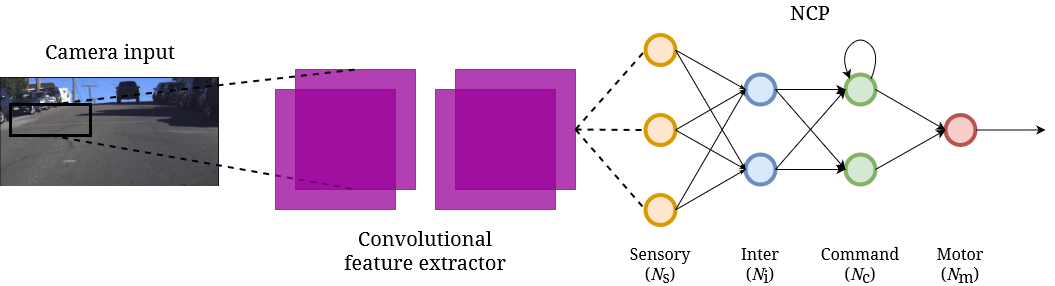}}\\ (a) \\
        \frame{\includegraphics[width=.75\linewidth]{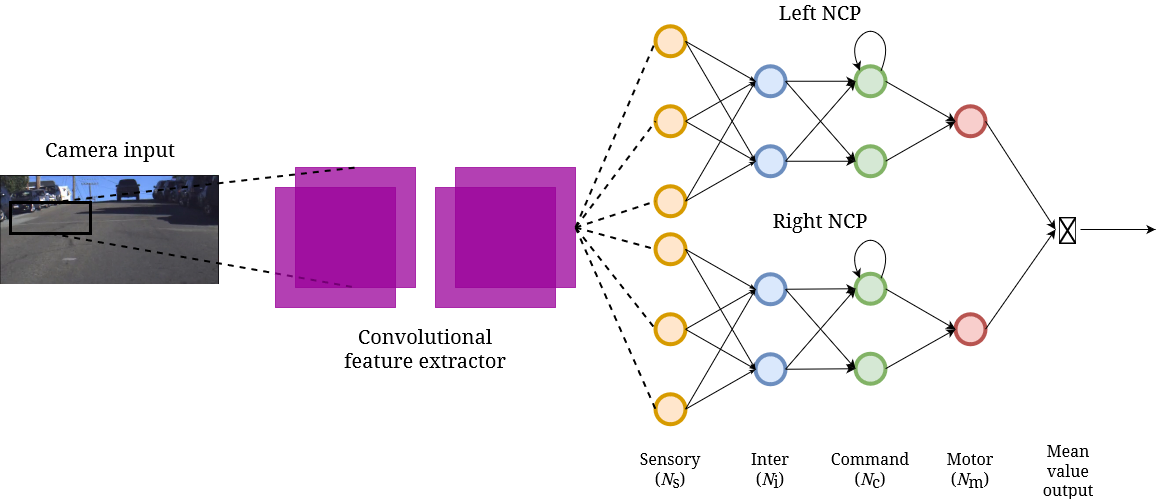}} \\ (b) \\  \begin{tabular}{|c|c|c|c|c|}
        \rowcolor{lightgray}
        \hline
        & \multicolumn{2}{c|}{\textbf{Left NCP}} & \multicolumn{2}{c|}{\textbf{Right NCP}}  \\
        \cline{2-5}
        \multirow{-2}{*}{\thead{\textbf{\qquad\, Model \qquad\,}}}& \textbf{Inter} & \textbf{Command} & \textbf{Inter} & \textbf{Command} \\
        \rowcolor{white}
        \hline
        CNN-DNCP & 3 & 5 & 4 & 6 \\
        \hline
        CNN-DNCP v2 & 9 & 7 & 12 & 8 \\
        \hline
        CNN-DNCP v3 & 12 & 8 & 5 & 3 \\
        \hline
        CNN-DNCP v4 & 12 & 8 & 5 & 3 \\
        \hline
        \end{tabular}
\\
        (c)
        \end{tabular}
        \caption{Generalised versions (not all model neurons are displayed) of the hybrid brain-inspired model architectures and their NCP wiring settings. The sensory and motor neurons are set to 100 and 1 respectively for all models: (a) CNN-NCP structure (inter = 12, command = 8); (b) CNN-DNCP structure; (c) NCP wiring configurations of the developed CNN-DNCP models.}
        \label{fig:binarch}
        \end{figure}

        The front-end layers of CNN mode is placed before brain-inspired module for beginning with preliminary image feature extraction. These visual attribute extractions are then passed to the brain-inspired RNN, which utilises them along with the hidden state for sequential feature extraction and to predict the corresponding steering angles one by one. Each returned steering angle is determined via the modified features and the hidden state, which keeps the details from previous predictions intact. For each newly developed model, we keep the chosen CNN feature engineering head as it is except the last one. Moreover, every model has different NCP wiring settings. To make different settings more distinguishable,  we name them by using a variant encoding up to fourth version (i.e. CNN-DNCP v4), as seen in Figure \ref{fig:binarch}. The wirings are sparsely designed using the NCP principles introduced in \cite{lechner}. Any instance of the LTC is essentially an RNN since it is defined as a series of ordinary differential equations in time.
        
    \section{Experimental Evaluation}
    
    \subsection{Datasets}
    All the developed models are evaluated on 
    Udacity Simulator and comma.ai datasets.
    
        \textbf{Udacity Simulator Dataset}: The initial dataset used to train the networks had been generated from Udacity's self-driving car simulator. The simulator has two modes — training and autonomous. The training mode is used to capture training data while the autonomous mode is used to assess a learned model by running a driving script that autonomously navigates the car on the given track. Also, we use the lakeside track as our training data whereas the jungle track is employed as a testing ground for the already trained models. The developed architectures are evaluated with no prior knowledge about the features of the map. The reason behind is to assess the generalisation ability of the models in a previously unseen environment.
        
        The Udacity dataset preprocessing includes: cropping and resizing the camera images from 160 x 320 to 66 x 200; colour space conversion from RGB to YUV; image data rescaling from [0, 255] to the [-1, 1] range. Moreover, we also perform data augmentation: choosing between the left, middle and right camera perspectives; image flipping; image shifting; adding shadows; brightness alteration.
        
        \textbf{Comma.ai Dataset}: The comma.ai driving dataset employed is a publicly released part of the driving data adopted by Santana et al. \cite{santana}. The video and sensors in the dataset are the same as those used in the comma.ai self-driving vehicle research platform. In the released dataset there is a total of 45 GB or 7.25 hours of driving data divided into 11 video clips of variable size. Further detail about the dataset and the measurement equipment used can be found online on the companion website \cite{commares}.
        
        During the data preprocessing step we observe the dataset recordings and split them into 3 - sunny, cloudy and night time camera recordings. The sunny weather videos are adopted for training while the other two types are left for architecture evaluation. In addition, the CNN-DNCP model architecture is dropped since it takes way too long to train and the loss result is strikingly unsatisfactory seen against those of the other models.

\subsection{Implementation Details}
We use Tensorflow with a single RTX 2060 GPU to train various networks. In this work, we split the data into 80\% training set and 20\% validation set to measure the performance after each epoch. We adopt the ReLU activation function. In most cases, activation functions are kept responsible for converting the sum of the node’s weighted input into either its activation or output. ReLU is a piecewise linear function that outputs zero if the input is negative and the input itself otherwise. For the loss function, we use MSE, which is the default loss function for a regression problem. In this work, we use the Adaptive Moment Estimation (Adam) as optimiser to minimise the loss function. The maximum number of epochs is set to 10 for training on above mentioned two datasets.

\subsection{Performance Comparison}
Two kinds of performance are evaluated in this paper, which are training performance and generalisation performance.

\begin{figure}
        \centering
        \begin{tabular}{cc}
        \includegraphics[width=5.5cm]{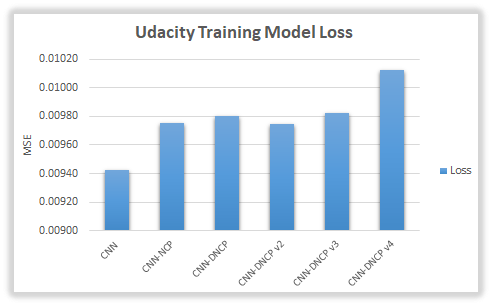}&
        \includegraphics[width=5.365cm]{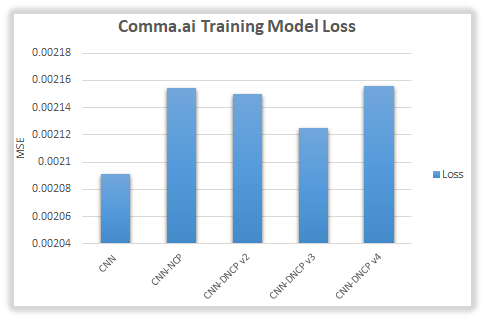}\\
        (a)&(b)
        \end{tabular}
        \caption{Model training performances (MSE results) on both datasets: (a) Udacity lakeside track; (b) Comma.ai sunny recordings.}
        \label{fig:training}
        \end{figure}

        \textbf{Training Performance}: In Figure \ref{fig:training}, we observe the training results of models on Udacity and Comma.ai datasets. Regarding the performance on Udacity dataset, the CNN model achieves an MSE result that is 3.28\% lower than the best CNN-DNCP brain-inspired model. In terms of loss performance on the comma.ai dataset, CNN once again outperforms the newly created brain-inspired models as seen in the figure. In other words, the CNN model delivers an MSE that is 1.6\% better than the CNN-DNCP v3 framework outcome. Please notice that these are training results. A model performing well on the data from the same source domain does not guarantee that the model can generalise well on the data from another domain. Therefore, the generalisation ability of the models is also evaluated in the following section.

\begin{figure}
        \centering
        \begin{tabular}{c}
        \includegraphics[width=5.5cm]{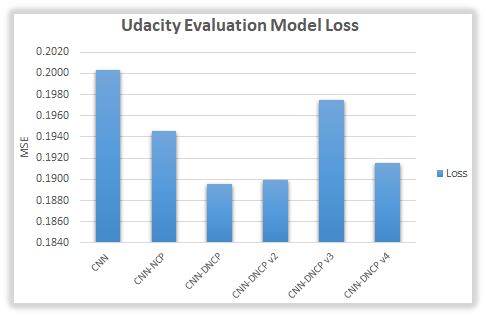}\\ 
        (a)
        \end{tabular}
        \begin{tabular}{cc}
        \includegraphics[width=5.365cm]{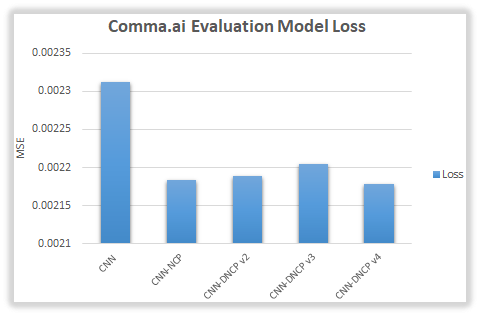}&
        \includegraphics[width=5.365cm]{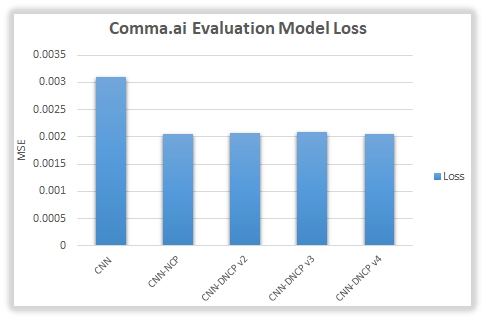}\\
        (b)&(c)
        \end{tabular}
        \caption{Model evaluation performances (MSE results) on both datasets: (a) Udacity jungle track; (b) Comma.ai cloudy weather recordings; (c) Comma.ai night time recordings.}
        \label{fig:evaluation}
        \end{figure}

    \textbf{Generalisation Performance}: with the training performance, the brain-inspired (CNN-DNCP v2) model can reduce 6.57\% error when comparing with CNN model on Udacity dataset,  which can be found in Figure \ref{fig:evaluation}.a. The models are trained on Udacity lakeside track data and evaluated on Udacity jungle map data for assessing the generalisation performance. Moreover, the Udacity simulator evaluation recordings are available at our \href{https://www.youtube.com/watch?v=C6ITvRRaEvQ&list=PLPqqcK1-S66E1MdPkFKbS66BwlhvFmpY_}{Youtube channel playlist}.  In Figure \ref{fig:evaluation}.b and \ref{fig:evaluation}.c, we evaluate the generalisation performance by using cloudy weather and night time data from Comma.ai dataset, where models are trained on a kind of weather condition and evaluated on another kind of weather condition. The brain-inspired architecture (CNN-DNCP v4) is 5.80\% and 34.05\% more effective than the CNN model with regard to MSE on the cloudy and night time comma.ai recordings correspondingly.According to the experimental results, we observe that the brain-inspired models perform significantly better than the CNN reference architecture in terms of MSE on generalisation performance.

\section{Conclusion and Future Work}

    In this paper, we propose a brain-inspired deep imitation learning method to enhance generalisation ability on autonomous driving. The design of the proposed network is inspired by neuroscientific evidence on the functional asymmetry of the two sides of the human brain. In particular, we develop 5 variants of brain-inspired networks, which combine visual feature extracted from CNN with temporal recursion networks (NCP) for imitating human driving behaviours. The developed networks are evaluated on two big autonomous driving datasets, including Udacity simulation dataset and comma.ai real-world dataset in terms of their generalisation ability. Extensive experiments indicate that the designs of brain-inspired network architectures, such as the CNN-NCP and CNN-DNCP, demonstrate promising performance on end-to-end autonomous driving vehicles. Moreover, by using stacked composite models (which communicate time-related and sensory characteristics in addition to constant-length input representations) an increased generalisation performance is achievable against conventional autonomous driving methods such as CNNs. There is still room for future work. Since the functions of layers of DL networks are insufficiently understood and the design of networks lacks theory support from psychology and neuroscience, the future work can be linking a well-characterised neural architecture of social learning in humans to DL networks, examining whether the application of this psychological theory can provide novel insights for improving the efficiency of DL networks by reducing redundant layers of learning. 

\section*{Acknowledgments}
This work was was supported by the University of Aberdeen Internal Funding to Pump-Prime Interdisciplinary Research and Impact under grant number SF10206-57.

%Bibliography
\bibliographystyle{unsrt}  
\bibliography{bio-inspired_IL}

\end{document}